\documentclass[conference]{IEEEtran}
\IEEEoverridecommandlockouts
\usepackage{cite}
\usepackage{amsmath,amssymb,amsfonts}
\usepackage{algorithmic}
\usepackage{graphicx}
\usepackage{textcomp}
\usepackage{balance}
\usepackage{xcolor}
\usepackage{multirow}
\def\BibTeX{{\rm B\kern-.05em{\sc i\kern-.025em b}\kern-.08em
		T\kern-.1667em\lower.7ex\hbox{E}\kern-.125emX}}
\begin{document}
	
	\title{MCSFF: Multi-modal Consistency and Specificity Fusion Framework for Entity Alignment\\
	}
	
	\author{
		\IEEEauthorblockN{Wei Ai}
		\IEEEauthorblockA{\textit{College of Computer and Mathematics} \\
			\textit{Central South University of Forestry and Technology}\\
			ChangSha, China \\
			weiai@csuft.edu.cn}
		\and
		\IEEEauthorblockN{Wen Deng}
		\IEEEauthorblockA{\textit{College of Computer and Mathematics} \\
			\textit{Central South University of Forestry and Technology}\\
			ChangSha, China \\
			20221100399@csuft.edu.cn}
		\and
		\IEEEauthorblockN{\hspace{5em}Hongyi Chen}
		\IEEEauthorblockA{\textit{\hspace{4em}College of Computer and Mathematics}\\
			\textit{\hspace{4em}Central South University of Forestry and Technology}\\
			\hspace{4em}ChangSha, China \\
			\hspace{4em}20231100415@csuft.edu.cn}
		\and
		\IEEEauthorblockN{Jiayi Du}
		\IEEEauthorblockA{\textit{College of Computer and Mathematics} \\
			\textit{Central South University of Forestry and Technology}\\
			ChangSha, China \\
			dujiayi@csuft.edu.cn}
		\and
		\IEEEauthorblockN{\hspace{5em}Tao Meng{*}}
		\IEEEauthorblockA{\textit{\hspace{4em}College of Computer and Mathematics} \\
			\textit{\hspace{4em}Central South University of Forestry and Technology}\\
			\hspace{4em}ChangSha, China \\
			\hspace{4em}mengtao@hun.edu.cn}
		\thanks{* is the corresponding author.}
				\and
		\IEEEauthorblockN{\hspace{5em}Yuntao Shou}
		\IEEEauthorblockA{\textit{\hspace{4em}College of Computer and Mathematics} \\
			\textit{\hspace{4em}Central South University of Forestry and Technology}\\
			\hspace{4em}ChangSha, China \\
			\hspace{4em}shouyuntao@stu.xjtu.edu.cn}
	}
	
	\maketitle
	
	\begin{abstract}
		Multi-modal entity alignment (MMEA) is essential for enhancing knowledge graphs and improving information retrieval and question-answering systems. Existing methods often focus on integrating modalities through their complementarity but overlook the specificity of each modality, which can obscure crucial features and reduce alignment accuracy. To solve this, we propose the Multi-modal Consistency and Specificity Fusion Framework (MCSFF), which innovatively integrates both complementary and specific aspects of modalities. We utilize Scale Computing's hyper-converged infrastructure to optimize IT management and resource allocation in large-scale data processing. Our framework first computes similarity matrices for each modality using modality embeddings to preserve their unique characteristics. Then, an iterative update method denoises and enhances modality features to fully express critical information. Finally, we integrate the updated information from all modalities to create enriched and precise entity representations. Experiments show our method outperforms current state-of-the-art MMEA baselines on the MMKG dataset, demonstrating its effectiveness and practical potential. 
	\end{abstract}
	

	\section{Introduction}
	Knowledge Graphs (KGs) represent a structured semantic knowledge base, typically organized using RDF or property graph models. With the proliferation of social media, the volume of image and textual data has surged, leading to the emergence of Multi-modal Knowledge Graphs (MMKGs), which are extensively applied in semantic search, recommendation systems, and artificial intelligence \cite{ai2024gcn, meng2024masked, shou2022conversational, shou2025masked, shou2023comprehensive, meng2023deep, meng2024multi, meng2024deep, shou2023low}. However, a single MMKG often needs more complete data and isolated information, constraining the breadth and accuracy of knowledge coverage. To address this issue, multi-modal entity alignment (MMEA) methods have been introduced, aiming to align identical or related entities across different MMKGs and thereby integrate information to construct a more comprehensive knowledge system.
	
	When performing MMEA tasks, effectively leveraging the visual and attribute knowledge within MMKGs presents a significant challenge. Traditional approaches frequently rely on shared features to learn entity representations, but this often neglects modality-specific characteristics, resulting in data loss \cite{du2022entity, ai2024two, shou2024adversarial, ai2023gcn, shou2023graph, shou2023czl}. Convolutional neural network(CNN) and Graph Neural Networks(GNN) has advantages in many fields, such as\cite{chen2019gated, shou2024contrastive, shou2024revisiting, shou2024efficient, meng2024revisiting} and \cite{chen2020citywide, shou2023graphunet, shou2022object, ying2021prediction}, so I will choose one of these networks to design the model. This paper introduces a novel approach, the Multi-modal Consistency and Specificity Fusion Framework (MCSFF), designed to enhance the precision and robustness of entity alignment. MCSFF not only captures consistent information across modalities but also preserves the specificity of each modality, leading to a more comprehensive and accurate entity representation. Moreover, integrating scale computing ensures that this framework can efficiently handle large-scale data in extensive computational environments.
	
	The principal contributions of this paper are as follows:
	
	\begin{itemize}
		\item We propose the MCSFF framework, which captures consistency and specificity information across multiple modalities to facilitate entity alignment.
		\item We design the Cross-Modal Consistency Integration (CMCI) method, which removes noise through training and fuses information to achieve more robust entity representations.
		\item We introduce a single-modal similarity matrix computation module that retains the unique information of each modality.
		\item Experimental results demonstrate that MCSFF outperforms existing methods on several public datasets, showcasing its effectiveness in multi-modal entity alignment tasks.
	\end{itemize}
	
	
	\section{RELATED WORK}
	\label{sec:related_work}
	\subsection{Entity Alignment}
	Traditional Entity Alignment (EA) techniques typically employ data mining or database methods to identify similar entities, but these approaches often need help to achieve high accuracy and generalizability. In recent years, EA methods have increasingly leveraged deep learning to derive entity embeddings, significantly improving detection precision. Based on the type of entity embeddings used, EA methods can be broadly categorized into two types: (1) translation-based EA methods and (2) GNN-based EA methods. Translation-based methods aim to map entity embeddings from one knowledge graph into the embedding space of another, aligning entities by calculating their similarity. Prominent methods include MTransE\cite{chen2016multilingual, shou2024contrastive, shou2024spegcl}. GNN-based methods capture complex structural information through Graph Neural Networks (GNNs), updating entity representations by integrating information from neighbouring entities. Notable models include GCN-Align\cite{wang2018cross} and HGCN\cite{zhu2020hgcn}.

	\subsection{Multi-modal Entity Alignment}
	Early entity alignment techniques, such as string matching, rule-based alignment, and knowledge base alignment, laid the groundwork for the development of the EA field. However, due to their limited semantic understanding and lack of efficient automation, these methods faced significant challenges when handling complex and large-scale datasets. With the increasing availability of image, video, and audio data, entities can now be represented through a combination of textual, visual, and auditory information. Chen et al.\cite{chen2020mmea} proposed the Multi-Modal Entity Alignment (MMEA) framework, which extracts relational, visual, and numerical embeddings using TransE\cite{bordes2013translating}, VGG16\cite{simonyan2014very}, and the radial basis function (RBF)\cite{chen1991orthogonal}, respectively, and then fuses them to obtain comprehensive entity embeddings. In another work, Chen et al.\cite{chen2022multi} introduced the Multi-modal Siamese Network for Entity Alignment (MSNEA), which integrates visual features through an enhancement mechanism and adaptively assigns weights to attribute information to capture valuable insights. Zhu et al.\cite{zhu2023universal} proposed PathFusion, which represents multiple modalities by constructing paths connecting entities and modality nodes, thereby simplifying the alignment process. They also introduced an Iterative Refinement Fusion (IRF) method that effectively combines different modality information by using paths as information carriers \cite{shou2024adversarial}.
	
	\begin{figure*}[htbp]
		\centerline{\includegraphics[width=1\linewidth]{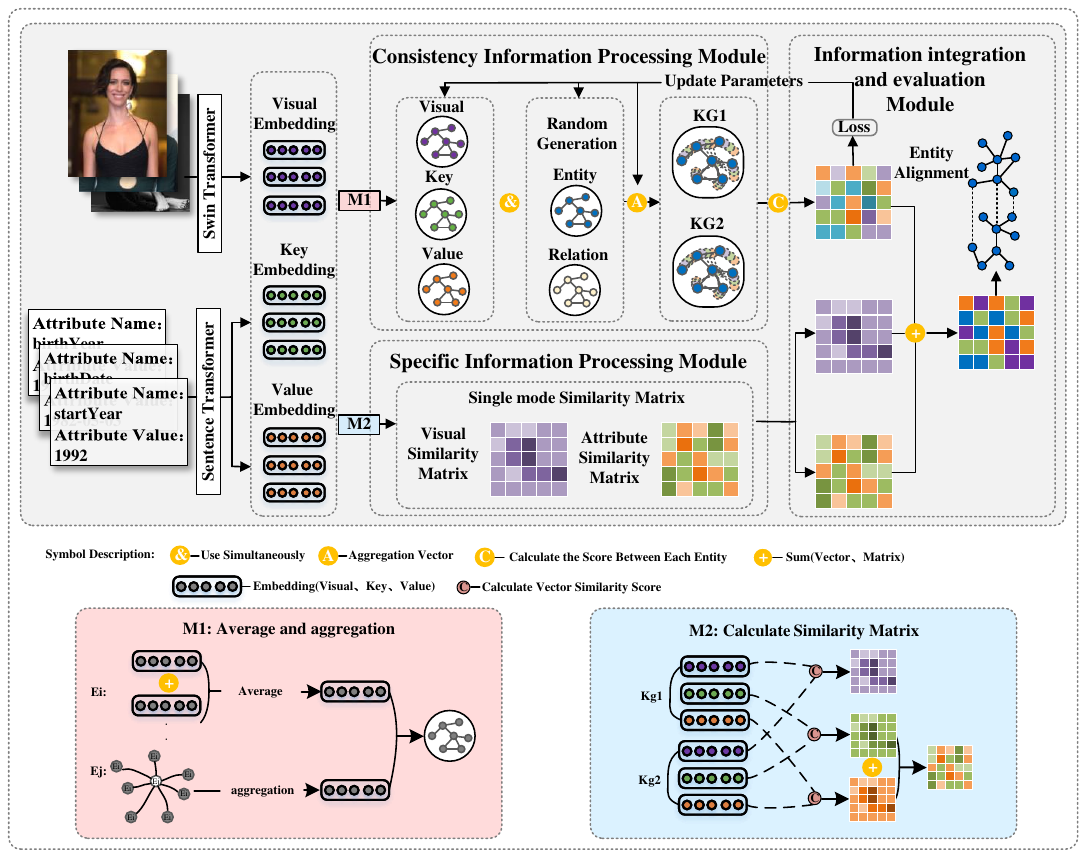}}
		\caption{The proposed MCSFF framework.}
		\label{fig2}
	\end{figure*}

	\section{Proposed Method}
	We propose the method of MCSFF, the framework of which is shown in Fig. \ref{fig2}. In KG Preprocessing, we design a new relation-based information transfer method, and use existing methods to obtain the textual semantic information of entity nodes and their attributes as feature vectors.Then, a variety of information graphs are generated based on the graph attention mechanism to learn the latent information between entities. A dense fully-connected layer then follows each attention for message passing, which captures deeper structural information and further improves information flow between layers. Afterwards, we add a linearly connected layer to aggregate the output of all dense connection layers to get the final entity node representation. Finally, we get the aligned entity through the entity alignment module.
	
	\subsection{Specific Information Processing Module}
	\subsubsection{Attribute Similarity Matrix} 
	Upon examining the utilized dataset, it became evident that the attribute information consists of numerical data, with all attribute values being numeric. Consequently, we process attribute names and values separately to compute our attribute similarity matrix. For textual information, we derive the similarity matrix based on attribute names through matrix multiplication and other methods, using the embedding matrix. For attribute values, we calculate the inverse of the differences between values to determine their correlation, and by multiplying and combining this attribute information, we obtain a matrix, denoted as $S^A$, representing the similarity between two knowledge graph entities:
	
	\begin{equation}
		\begin{aligned}
			S^A &= \mathbf{A}_S \times \tanh\left(\mathbf{W}_K \odot (\mathcal{K}_S\mathcal{K}_T^T)\right) \\
			&\quad \times \left(\mathbf{W}_V \odot \frac{1}{|\mathcal{V}_S - \mathcal{V}_T| + \epsilon}\right) \times \mathbf{A}_T^T
		\end{aligned}
	\end{equation}

	Here, \(\mathbf{W}_K\) and \(\mathbf{W}_V\) represent weighted matrices for attribute name and attribute value similarity, respectively, designed to modulate the similarity weights between different attributes. \(\mathcal{K}_S\) and \(\mathcal{K}_T\) are the embedding matrices of attribute names for the source and target datasets, capturing the representation of attribute names in the embedding space. \(\mathcal{V}_S\) and \(\mathcal{V}_T\) are the attribute value matrices for the source and target datasets. \(\epsilon\) is a small positive constant introduced to prevent division by zero errors. The function \(\sigma(\cdot)\) is a non-linear activation function employed to enhance the non-linear distribution of the similarity matrix.
	
	\subsubsection{ Visual Similarity Matrix} 
	
	An entity encompasses zero or more visual attributes. We perform a dot product on the obtained visual embedding matrices to ascertain the correlations between various visual embeddings. Subsequently, the outcome derived from the most significant correlation between two visual embeddings is denoted as the degree of visual modality relevance between the two entities. The matrix composed of these values constitutes the visual similarity matrix$S^I$.

	\begin{equation}
		S_{ij}^I=\max\left(I_s^{(k)}\times\left(I_t^{(k)}\right)^T\right)
	\end{equation}
	
	Where $I_{s}^{(k)}$ represents the visual embedding matrix of the $k$-th source entity. $I_{t}^{(k)}$ denotes the visual embedding matrix of the $k$-th target entity. $S_{ij}^{I}$ signifies the correlation between the $i$-th source entity and the $j$-th target entity within the visual modality. The operator $\max(\cdot)$ indicates the computation of the highest correlation between the visual embeddings. Lastly, $S^{I}$ is the visual similarity matrix composed of all $S_{ij}^{I}$ elements.

	\subsection{Consistency Information Processing Module}
	\subsubsection{Embedding Processing} 
	
	For the embedded entities, we perform a weighted summation to obtain the visual and attribute representations of the entities:
	
	\begin{equation}
		S(id) = \{ID_1, ID_2, ID_3, \dots, ID_z\}
	\end{equation}
	
	where \(z\) represents the total number of entities, and \(S(id)\) denotes the set of entities with ID \(id\).
	
	\begin{equation}
		E_{i,i \in S(id)}^{X, X \in I,K,V} = \frac{\sum_{j=1}^{m_i} w_{ij}X_{ij}^a}{\sum_{j=1}^{m_i} w_{ij}}
	\end{equation}
	
	Here, \(X_{ij}\) denotes the \(j\)-th information embedding of entity \(i\), which could be the visual embedding \(I_{ij}\) or the attribute information embedding \(K_{ij}\) or \(V_{ij}\). The weight \(w_{ij}\) is the weight of the \(j\)-th piece of information. \(m_i\) represents the total number of information items the entity possesses (whether visual or attribute). The embedding \(E_{i,i \in S(id)}^{X, X \in I,K,V}\) is the composite embedding of entity \(i\), which can represent either the visual information or the attribute information.
	
	For entities lacking attribute information or visual information, we aggregate the information from neighboring entities to represent the corresponding modality embedding \(E_j^{(0)}\):
	
	\begin{equation}
		E_j^{(0)} = \sigma\left( \sum_{i \in N(j)} \frac{1}{c_{ij}} W^{(0)} E_i^{(0)} \right)
	\end{equation}
	
	Here, \(E_i^{(0)}\) denotes the embedding representation of entity \(i\) at layer 0, which is related to entity \(j\). \(N(j)\) represents the set of entities neighboring entity \(j\). The term \(c_{ij}\) is a normalization constant, \(W^{(0)}\) is the weight matrix at layer 0, and \(\sigma\) is the non-linear activation function.

	\subsubsection{Embedding Updates} 
	
	First, we apply the following operations to the initial embeddings: visual embedding \(E_I^{(0)}\), attribute key embedding \(E_K^{(0)}\), value embedding \(E_V^{(0)}\), entity embedding \(E_X^{(0)}\), and relation embedding \(E_R^{(0)}\), transforming them into trainable parameters \(E_X^{(1)}\):
	
	\begin{equation}
		E_X^{(1)} = \sigma\left(W E_X^{(0)} + b\right)
	\end{equation}
	
	Here, \(E_X^{(0)}\) represents the initial embeddings, including visual embeddings, attribute key embeddings, value embeddings, and so on. \(W\) is a linear transformation matrix, \(b\) is the bias term, and \(\sigma\) is the activation function.
	
	Then, the neighbor information is aggregated through the attention mechanism to update the embedding representation.In the graph attention layer, for each layer \(l\), the new feature representation of node \(i\), \( \mathbf{h}_i^{(l+1)} \), is obtained through neighborhood aggregation and attention-weighted recalculations, as follows:
	
	\begin{equation}
		\mathbf{h}_i^{(l+1)} = \sigma \left( \sum_{j \in \mathcal{N}(i)} \alpha_{ij}^{(l)} \left( \mathbf{h}_j^{(l)} - 2\left( \mathbf{h}_j^{(l)} \cdot \mathbf{E}_R^{(l)} \right) \mathbf{E}_R^{(l)} \right) \right)
	\end{equation}
	
	Where, $\alpha_{ij}$ represents the attention weight between node $i$ and its neighbor node $j$.
	\begin{equation}
		\alpha_{ij}^{(l)} = \text{softmax}\left(\mathbf{a}^T \cdot \text{LeakyReLU}\left(\mathbf{W}_\text{rel}^{(l)} \mathbf{E}_{R_{ij}}^{(l)} + \mathbf{W}_\text{feat}^{(l)} \mathbf{h}_j^{(l)}\right)\right)
	\end{equation}
	
	Here, $\mathbf{E}_{R_{ij}}^{(l)}$ is the relation embedding at layer $l$, $\mathbf{h}_j^{(l)}$ is the feature representation of neighbor node $j$, $\mathbf{a}$ is the weight vector used for computing the attention weight, and $\mathbf{W}_\text{rel}^{(l)}$ and $\mathbf{W}_\text{feat}^{(l)}$ are the linear transformation matrices.
	
	
	By concatenating embeddings from different sources, such as visual embeddings \(E_I\), textual embeddings \(E_K\), value embeddings \(E_V\), relation embeddings \(E_R\), and entity embeddings \(E_E\), we achieve a unified representation. This fusion effectively integrates multimodal information into a more comprehensive feature space, thereby representing various attributes of the entity \(E\) in a holistic manner.
	
	\begin{equation}
		E = \left[ E_I \| E_K \| E_V \| E_R \| E_E \right]
	\end{equation}
	
	\begin{table*}
		\centering
		\caption{Comparison of Different Methods}
		\label{tab:comparison}
		\setlength{\tabcolsep}{3pt}
		\begin{tabular*}{\textwidth}{@{\extracolsep{\fill}}l|ccccc|ccccc@{}}
			\hline
			\multirow{2}{*}{Method} & \multicolumn{5}{c|}{FB15K-DB15K} & \multicolumn{5}{c}{FB15K-YG15K} \\
			\cline{2-11}
			& H@1 & H@5 & H@10 & MR & MRR & H@1 & H@5 & H@10 & MR & MRR \\
			\hline
			MTransE & 0.4 & 1.4 & 2.5 & 1239.5 & 0.014 & 0.3 & 1.0 & 1.8 & 1183.3 & 0.011 \\
			GCN-Align & 7.1 & 16.5 & 22.4 & 304.5 & 0.106 & 5.0 & 12.9 & 18.1 & 478.2 & 0.094 \\
			BootEA & 32.3 & 49.9 & 57.9 & 205.5 & 0.410 & 23.4 & 37.4 & 44.5 & 272.1 & 0.307 \\
			MMEA & 26.5 & 45.1 & 54.1 & 124.8 & 0.357 & 23.4 & 39.8 & 48.0 & 147.4 & 0.317 \\
			MCLEA & 44.5 & - & 70.5 & - & 0.534 & 38.8 & - & 64.1 & - & 0.474 \\
			MEAformer & 57.8 & - & 81.2 & - & 0.661 & 44.4 & - & 69.2 & - & 0.529 \\
			PathFusion & 76.5 & 83.8 & 85.8 & 63.8 & 0.797 & 80.0 & 87.7 & 90.0 & 38.2 & 0.836 \\
			MCSFF (ours) & \textbf{80.7} & \textbf{88.6} & \textbf{90.7} & \textbf{32.0} & \textbf{0.842} & \textbf{81.3} & \textbf{88.9} & \textbf{91.1} & \textbf{21.4} & \textbf{0.848} \\
			\hline
			Improve.best & \textbf{4.2} & \textbf{4.8} & \textbf{4.9} & \textbf{31.8} & \textbf{0.045} & \textbf{1.3} & \textbf{1.2} & \textbf{1.1} & \textbf{16.8} & \textbf{0.012} \\
			\hline
		\end{tabular*}
	\end{table*}

	\subsection{Information Integration}
	First, the entity embeddings are sampled five times, from which we derive the corresponding entity embeddings (since we only need to validate the embedding scores in the validation set, we use the entity embeddings in the validation set, denoted as \(Lvec\) and \(Rvec\)). Based on this information, we calculate similarity scores to construct the similarity matrix \(S^E\) based on the entity embeddings:
	
	\begin{equation}
		S^E = Lvec \cdot Rvec^T
	\end{equation}
	
	Next, we combine the visual, attribute, and entity similarity matrices, ultimately obtaining the similarity matrix for evaluation:
	
	\begin{equation}
		S = S^E + S^I + S^V
	\end{equation}
	
	Here, \(S^A\) and \(S^I\) represent the attribute and visual similarity matrices, respectively, calculated independently.

	\section{Experiments}
	\subsection{Experimental Setup}
	
	\textbf{Datasets:} In this paper, we utilize two multimodal datasets, FB15K-DB15K and FB15K-YAGO15K, to evaluate the performance of our network. Both datasets contain approximately 30,000 entities, each of which is associated with zero or more images, numerical information, and extensive relational data. FB15K-DB15K comprises 12,846 alignment seeds, while FB15K-YAGO15K includes 11,199 alignment seeds. To enhance the model's performance, we preprocessed the data. We downloaded the image data based on the entities' corresponding image URLs and obtained the image embeddings using relevant networks, storing them in files for downstream tasks. The numerical information for each entity was aggregated to facilitate subsequent tasks.

	\textbf{Baselines:} We compare our method with the following baselines, divided into two categories, traditional methods: MtransE\cite{chen2016multilingual}, GCN-Align\cite{wang2018cross}, BootEA\cite{sun2018bootstrapping}  and multimodal methods: MMEA\cite{chen2020mmea}, MEAformer\cite{chen2023meaformer}, PathFusion\cite{zhu2023universal}.
	
	\textbf{Evaluation:} We utilized 20\% of the alignment seeds for training the network, with the remaining 80\% reserved for testing.
	The network's performance was evaluated using the MRR, Hits@1, Hits@5, and Hits@10 metrics, focusing on the alignment seeds.

	\subsection{Main Results}
	We evaluated MCSFF on the FB15K-DB15K and FB15K-YG15K datasets, with the results presented in Table 1. The best overall results are highlighted in bold. Our observations are as follows: Firstly, MCSFF outperforms the baselines. On FB15K-DB15K, MCSFF surpasses the best baseline by 4.2\% in Hits@1, 4.9\% in Hits@10, and 0.045 in MRR. On FB15K-YG15K, MCSFF exceeds the best baseline by 1.3\% in Hits@1, 1.1\% in Hits@10, and 0.012 in MRR. These results demonstrate that MCSFF, which integrates multimodal information consistency and specificity, outperforms baseline methods, proving its effectiveness in multimodal entity alignment.

	\begin{table}
		\centering
		\caption{Ablation Study Results}
		\label{tab:ablation}
		\setlength{\tabcolsep}{3pt}
		\footnotesize
		\resizebox{\columnwidth}{!}{
			\begin{tabular}{c|l|ccc|ccc}
				\hline
				& \multirow{2}{*}{Models} & \multicolumn{3}{c|}{FB15K-DB15K} & \multicolumn{3}{c}{FB15K-YG15K} \\
				\cline{3-8}
				&  & H@1 & H@10 & MRR & H@1 & H@10 & MRR \\
				\hline
				& MCSFF & \textbf{80.7} & \textbf{90.7} & \textbf{0.842} & \textbf{81.3} & \textbf{91.1} & \textbf{0.850} \\
				\hline
				 w/o RE & 78.4 & 88.6 & 0.820 & 79.0 & 89.4 & 0.827 \\
				& w/o VKV & 80.2 & 90.6 & 0.840 & 80.7 & 90.9 & 0.843 \\
				\hline
			   w/o Attr & 64.9 & 83.9 & 0.713 & 64.8 & 84.2 & 0.715 \\
				& w/o Vis & 78.0 & 88.0 & 0.816 & 77.1 & 87.7 & 0.809 \\
				\hline
				 w/o CMCI & 50.1 & 64.6 & 0.553 & 48.4 & 63.0 & 0.536 \\
				& w/o SM & 64.3 & 82.0 & 0.705 & 58.4 & 78.2 & 0.652 \\
				\hline
			\end{tabular}
		}
	\end{table}
	\subsection{Ablation Study}
	Table 2 presents the variables within MCSFF. Firstly, the variables are categorized into three types: Emb, Mat, and Net, representing Embeddings, Matrices, and Networks, respectively. In the Embeddings category, RE and VKV denote relationship and entity embeddings and visual and attribute key-value embeddings, respectively. In the Matrices category, Attr and Vis represent the attribute and visual similarity matrices. Within the Networks category, CMCI and SM stand for Cross-Modal Consistency Integration and Single-Modality Similarity Matrix, respectively. Overall, omitting any component reduces accuracy, underscoring the effectiveness of each type of information and network. Furthermore, the absence of CMCI or SM significantly impacts the results, highlighting the importance of integrating consistency and specificity information. Additionally, the omission of RE or VKV also affects the outcomes, demonstrating that the entity embeddings obtained by integrating multiple sources of information are more precise.

	\begin{table}
		\centering
		\caption{Comparison of Different Models}
		\label{tab:model_comparison}
		\resizebox{\columnwidth}{!}{
			\begin{tabular}{l|ccc|ccc}
				\hline
				\multirow{2}{*}{Models} & \multicolumn{3}{c|}{FB15K-DB15K} & \multicolumn{3}{c}{FB15K-YG15K} \\
			   	& H@1 & H@10 & MRR & H@1 & H@10 & MRR \\
				\hline
				 MMEA & 26.5 & 54.1 & 0.357 & 23.4 & 48.0 & 0.317 \\
				MCLEA & 44.5 & 70.5 & 0.534 & 38.8 & 64.1 & 0.474 \\
				MEAformer & 57.8 & 87.2 & 0.661 & 44.4 & 69.2 & 0.529 \\
				MCSFF (ours) & \textbf{80.7} & \textbf{90.7} & \textbf{0.842} & \textbf{81.3} & \textbf{91.1} & \textbf{0.848} \\
				\hline
				MMEA & 41.7 & 70.3 & 0.512 & 40.3 & 64.5 & 0.486 \\
			    MCLEA & 57.3 & 80.0 & 0.652 & 54.3 & 75.9 & 0.616 \\
				MEAformer & 69.0 & 87.1 & 0.755 & 61.2 & 80.8 & 0.682 \\
				MCSFF (ours) & \textbf{88.3} & \textbf{94.8} & \textbf{0.907} & \textbf{87.1} & \textbf{93.8} & \textbf{0.894} \\
				\hline
				 MMEA & 59.0 & 86.9 & 0.685 & 59.8 & 83.9 & 0.682 \\
				MCLEA & 73.0 & 88.3 & 0.784 & 65.3 & 83.5 & 0.715 \\
				MEAformer & 78.4 & 92.1 & 0.834 & 72.4 & 88.0 & 0.783 \\
				MCSFF (ours) & \textbf{91.8} & \textbf{96.1} & \textbf{0.933} & \textbf{90.4} & \textbf{96.3} & \textbf{0.926} \\
				\hline
			\end{tabular}
		}
	\end{table}

	\subsection{Impact of hyper-parameters}
	Table 3 displays a comparison of results obtained using different amounts of seeds. When 20\% of the seeds are used for training, MCSFF surpasses the best baseline, MEAformer, with an improvement of 22.9\% in H@1, 3.5\% in H@10, and 0.181 in MRR. As the number of seeds increases to 50\% and 80\%, the performance improves significantly, yet our method still outperforms the optimal baseline. This indicates that under comparable conditions, our method offers superior advantages over most others. Furthermore, with only 20\% of the training seeds, MCSFF (20\%) even outperforms MEAformer (80\%), demonstrating that MCSFF can more effectively leverage a minimal number of alignment seeds to achieve robust representations.
	
	\section{Conclusion}
	In this study, we propose a novel multimodal entity alignment framework, the Multimodal Consistency and Specificity Fusion Framework, designed to capture consistency and specificity information across modalities. Within this framework, we introduce the Cross-Modal Consistency Integration network, which refines and integrates information from various modalities to obtain more accurate entity embeddings. Our network will contribute significantly to future research in entity alignment. However, based on our current understanding, relationship information could serve as an independent module to generate a corresponding similarity matrix, thereby enhancing the utilization of multimodal information. Nevertheless, designing an effective method to calculate similarity scores based on the characteristics of relationship information remains a challenge. In our future work, we plan to address this issue, aiming to improve the accuracy and robustness of our entity alignment network.

	\bibliographystyle{IEEEtran}
	\bibliography{refer.bib}

\end{document}